\documentclass[conference]{IEEEtran}
\usepackage{times}

\usepackage[numbers]{natbib}
\usepackage{multicol}
\usepackage[bookmarks=true]{hyperref}
\usepackage{color}
\usepackage{graphicx}
\usepackage{amsmath}
\usepackage{amssymb}
\usepackage{subfloat}
\usepackage{multicol}

\newcommand{\sloss}{\textsc{SLoss}}
\newcommand{\pl}{\textsc{PLoss}}

\begin{document}

\title{PoseCNN: A Convolutional Neural Network for 6D Object Pose Estimation in Cluttered Scenes}

% You will get a Paper-ID when submitting a pdf file to the conference system
% \author{Author Names Omitted for Anonymous Review. Paper-ID [74]}

\author{\authorblockN{Yu Xiang$^{1,2}$, Tanner Schmidt$^2$, Venkatraman Narayanan$^3$ and Dieter Fox$^{1,2}$}
\authorblockA{$^1$NVIDIA Research, $^2$University of Washington, $^3$Carnegie Mellon University \\
yux@nvidia.com, tws10@cs.washington.edu, venkatraman@cs.cmu.edu, dieterf@nvidia.com}}

%\author{\authorblockN{Michael Shell}
%\authorblockA{School of Electrical and\\Computer Engineering\\
%Georgia Institute of Technology\\
%Atlanta, Georgia 30332--0250\\
%Email: mshell@ece.gatech.edu}
%\and
%\authorblockA{Twentieth Century Fox\\
%\and
%\authorblockN{James Kirk\\ and Montgomery Scott}
%\authorblockA{Starfleet Academy\\
%San Francisco, California 96678-2391\\

% avoiding spaces at the end of the author lines is not a problem with
% conference papers because we don't use \thanks or \IEEEmembership

%Homer Simpson\authorrefmark{2},
%Montgomery Scott\authorrefmark{3} and
%Eldon Tyrell\authorrefmark{4}}
%Telephone: (800) 555--1212, Fax: (888) 555--1212}
%\authorblockA{\authorrefmark{4}Tyrell Inc., 123 Replicant Street, Los Angeles, California 90210--4321}}

\maketitle

\begin{abstract}
	
Estimating the 6D pose of known objects is important for robots to interact with the real world. The problem is challenging due to the variety of objects as well as the complexity of a scene caused by clutter and occlusions between objects. In this work, we introduce PoseCNN, a new Convolutional Neural Network for 6D object pose estimation. PoseCNN estimates the 3D translation of an object by localizing its center in the image and predicting its distance from the camera. The 3D rotation of the object is estimated by regressing to a quaternion representation. We also introduce a novel loss function that enables PoseCNN to handle symmetric objects. In addition, we contribute a large scale video dataset for 6D object pose estimation named the YCB-Video dataset. Our dataset provides accurate 6D poses of 21 objects from the YCB dataset observed in 92 videos with 133,827 frames. We conduct extensive experiments on our YCB-Video dataset and the OccludedLINEMOD dataset to show that PoseCNN is highly robust to occlusions, can handle symmetric objects, and provide accurate pose estimation using only color images as input. When using depth data to further refine the poses, our approach achieves state-of-the-art results on the challenging OccludedLINEMOD dataset. Our code and dataset are available at \url{https://rse-lab.cs.washington.edu/projects/posecnn/}.

\end{abstract}

\IEEEpeerreviewmaketitle

\section{INTRODUCTION}

Recognizing objects and estimating their poses in 3D has a wide range of applications in robotic tasks. For instance, recognizing the 3D location and orientation of objects is important for robot manipulation. It is also useful in human-robot interaction tasks such as learning from demonstration. However, the problem is challenging due to the variety of objects in the real world. They have different 3D shapes, and their appearances on images are affected by lighting conditions, clutter in the scene and occlusions between objects.

Traditionally, the problem of 6D object pose estimation is tackled by matching feature points between 3D models and images \cite{lowe1999object,rothganger20063d,collet2011moped}. However, these methods require that there are rich textures on the objects in order to detect feature points for matching. As a result, they are unable to handle texture-less objects. With the emergence of depth cameras, several methods have been proposed for recognizing texture-less objects using RGB-D data \cite{hinterstoisser2012model,brachmann2014learning,Bo14Lea,schwarz2015rgb,kehl2016deep}. For template-based methods \cite{hinterstoisser2012model,hinterstoisser2012gradient}, occlusions significantly reduce the recognition performance. Alternatively, methods that perform learning to regress image pixels to 3D object coordinates in order to establish the 2D-3D correspondences for 6D pose estimation \cite{brachmann2014learning,brachmann2016uncertainty} cannot handle symmetric objects.

\begin{figure}
	\centering
	\includegraphics[height = 0.5\linewidth, width = \linewidth]{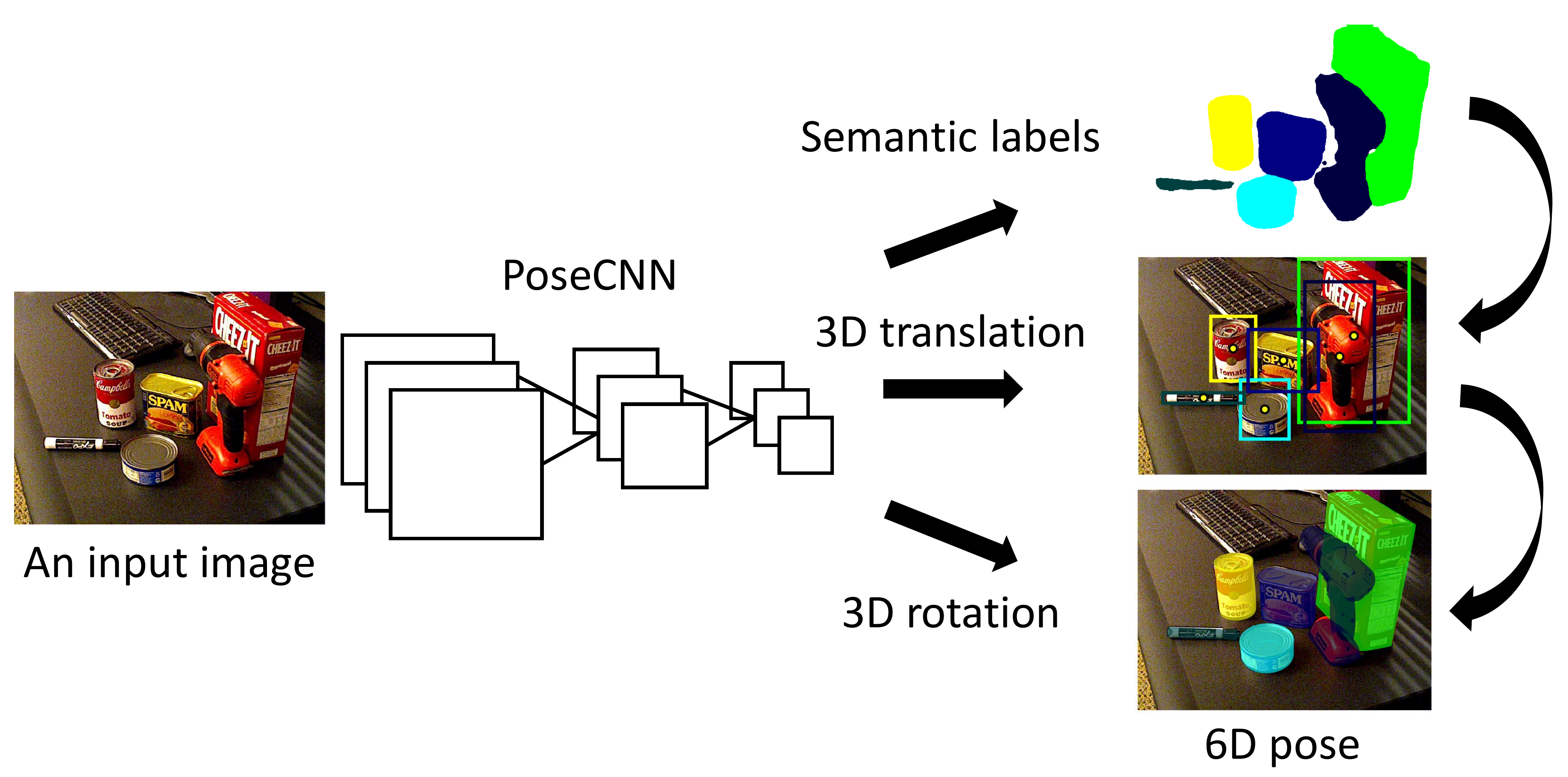}
	\caption{We propose a novel PoseCNN for 6D object pose estimation, where the network is trained to perform three tasks: semantic labeling, 3D translation estimation, and 3D rotation regression.}
	\label{fig:framework}
	\vspace{-8mm}
\end{figure}

In this work, we propose a generic framework for 6D object pose estimation where we attempt to overcome the limitations of existing methods. We introduce a novel Convolutional Neural Network (CNN) for end-to-end 6D pose estimation named PoseCNN. A key idea behind PoseCNN is to decouple the pose estimation task into different components, which enables the network to explicitly model the dependencies and independencies between them. Specifically, PoseCNN performs three related tasks as illustrated in Fig.~\ref{fig:framework}. First, it predicts an object label for each pixel in the input image. Second, it estimates the 2D pixel coordinates of the object center by predicting a unit vector from each pixel towards the center. Using the semantic labels, image pixels associated with an object vote on the object center location in the image. In addition, the network also estimates the distance of the object center. Assuming known camera intrinsics, estimation of the 2D object center and its distance enables us to recover its 3D translation $\mathbf{T}$. Finally, the 3D Rotation $\mathbf{R}$ is estimated by regressing convolutional features extracted inside the bounding box of the object to a quaternion representation of $\mathbf{R}$. As we will show, the 2D center voting followed by rotation regression to estimate $\mathbf{R}$ and $\mathbf{T}$ can be applied to textured/texture-less objects and is robust to occlusions since the network is trained to vote on object centers even when they are occluded.

Handling symmetric objects is another challenge for pose estimation, since different object orientations may generate identical observations. For instance, it is not possible to uniquely estimate the orientation of the red bowl or the wood block shown in Fig.~\ref{fig:ycb}. While pose benchmark datasets such as the OccludedLINEMOD dataset \cite{krull2015learning} consider a special symmetric evaluation for such objects, symmetries are typically ignored during network training. However, this can result in bad training performance since a network receives inconsistent loss signals, such as a high loss on an object orientation even though the estimation from the network is correct with respect to the symmetry of the object. Inspired by this observation, we introduce ShapeMatch-Loss, a new loss function that focuses on matching the 3D shape of an object. We will show that this loss function produces superior estimation for objects with shape symmetries.

We evaluate our method on the OccludedLINEMOD dataset~\cite{krull2015learning}, a benchmark dataset for 6D pose estimation.  On this challenging dataset, PoseCNN achieves state-of-the-art results for both color only  and RGB-D pose estimation (we use depth images in the Iterative Closest Point (ICP) algorithm for pose refinement). To thoroughly evaluate our method, we additionally collected a large scale RGB-D video dataset named YCB-Video, which contains 6D poses of 21 objects from the YCB object set~\cite{calli2015ycb} in 92 videos with a total of 133,827 frames. Objects in the dataset exhibit different symmetries and are arranged in various poses and spatial configurations, generating severe occlusions between them.

In summary, our work has the following key contributions:
\begin{itemize}
    \item We propose a novel convolutional neural network for 6D object pose estimation named PoseCNN. Our network achieves end-to-end 6D pose estimation and is very robust to occlusions between objects.
    \item We introduce ShapeMatch-Loss, a new training loss function for pose estimation of symmetric objects.
    \item We contribute a large scale RGB-D video dataset for 6D object pose estimation, where we provide 6D pose annotations for 21 YCB objects.
\end{itemize}

\noindent This paper is organized as follows. After discussing related work, we introduce PoseCNN for 6D object pose estimation, followed by experimental results and a conclusion.

\section{RELATED WORK}

6D object pose estimation methods in the literature can be roughly classified into template-based methods and feature-based methods. In template-based methods, a rigid template is constructed and used to scan different locations in the input image. At each location, a similarity score is computed, and the best match is obtained by comparing these similarity scores \cite{hinterstoisser2012gradient,hinterstoisser2012model,cao2016real}. In 6D pose estimation, a template is usually obtained by rendering the corresponding 3D model. Recently, 2D object detection methods are used as template matching and augmented for 6D pose estimation, especially with deep learning-based object detectors \cite{su2015render,rad2017bb8,kehl2017ssd,tekin2017real}. Template-based methods are useful in detecting texture-less objects. However, they cannot handle occlusions between objects very well, since the template will have low similarity score if the object is occluded.

In feature-based methods, local features are extracted from either points of interest or every pixel in the image and matched to features on the 3D models to establish the 2D-3D correspondences, from which 6D poses can be recovered \cite{lowe1999object,rothganger20063d,tulsiani2015viewpoints,pavlakos2017}. Feature-based methods are able to handle occlusions between objects. However, they require sufficient textures on the objects in order to compute the local features. To deal with texture-less objects, several methods are proposed to learn feature descriptors using machine learning techniques \cite{wohlhart2015learning,doumanoglou2016siamese}. A few approaches have been proposed to directly regress to 3D object coordinate location for each pixel to establish the 2D-3D correspondences \cite{brachmann2014learning,krull2015learning,brachmann2016uncertainty}. But 3D coordinate regression encounters ambiguities in dealing with symmetric objects.

In this work, we combine the advantages of both template-based methods and feature-based methods in a deep learning framework, where the network combines bottom-up pixel-wise labeling with top-down object pose regression. Recently, the 6D object pose estimation problem has received more attention thanks to the competition in the Amazon Picking Challenge (APC). Several datasets and approaches have been introduced for the specific setting in the APC \cite{rennie2016dataset,zeng2017multi}. Our network has the potential to be applied to the APC setting as long as the appropriate training data is provided.

\section{PoseCNN}

\begin{figure*}
	\centering
	\includegraphics[height = 0.58\linewidth, width = \linewidth]{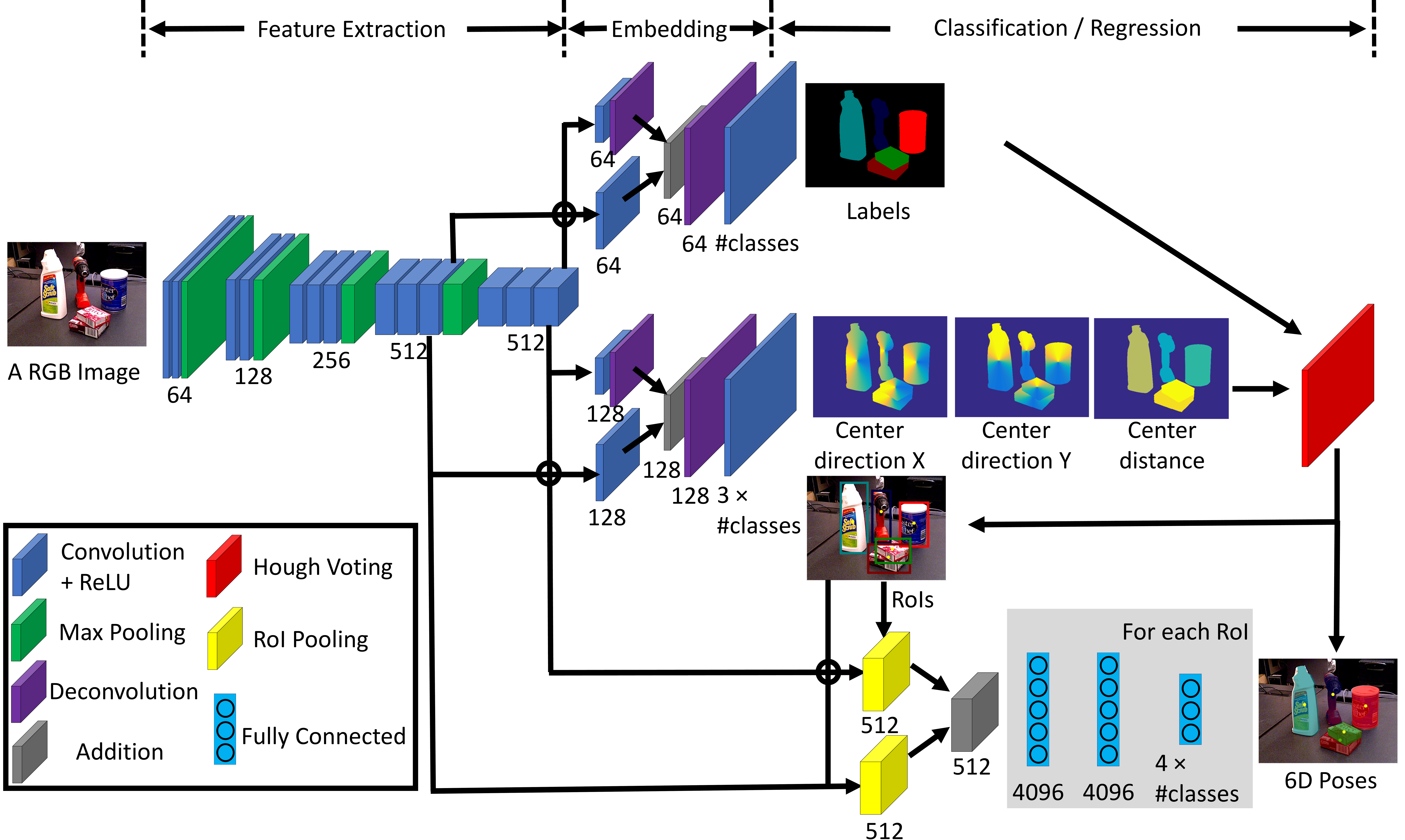}
	\caption{Architecture of PoseCNN for 6D object pose estimation.}
	\label{fig:network}
	\vspace{-4mm}
\end{figure*}

Given an input image, the task of 6D object pose estimation is to estimate the rigid transformation from the object coordinate system $O$ to the camera coordinate system $C$. We assume that the 3D model of the object is available and the object coordinate system is defined in the 3D space of the model. The rigid transformation here consists of an SE(3) transform containing a 3D rotation $\mathbf{R}$ and a 3D translation $\mathbf{T}$, where $\mathbf{R}$ specifies the rotation angles around the $X$-axis, $Y$-axis and $Z$-axis of the object coordinate system $O$, and $\mathbf{T}$ is the coordinate of the origin of $O$ in the camera coordinate system $C$.
In the imaging process, $\mathbf{T}$ determines the object location and scale in the image, while $\mathbf{R}$ affects the image appearance of the object according to the 3D shape and texture of the object. Since these two parameters have distinct visual properties, we propose a convolutional neural network architecture that internally decouples the estimation of $\mathbf{R}$ and $\mathbf{T}$.

\subsection{Overview of the Network}

Fig.~\ref{fig:network} illustrates the architecture of our network for 6D object pose estimation. The network contains two stages. The first stage consists of 13 convolutional layers and 4 max-pooling layers, which extract feature maps with different resolutions from the input image. This stage is the backbone of the network since the extracted features are shared across all the tasks performed by the network. The second stage consists of an embedding step that embeds the high-dimensional feature maps generated by the first stage into low-dimensional,  task-specific features. Then, the network performs three different tasks that lead to the 6D pose estimation, i.e., semantic labeling, 3D translation estimation, and 3D rotation regression, as described next.

\subsection{Semantic Labeling}

In order to detect objects in images, we resort to semantic labeling, where the network classifies each image pixel into an object class. Compared to recent 6D pose estimation methods that resort to object detection with bounding boxes \cite{rad2017bb8,kehl2017ssd,tekin2017real}, semantic labeling provides richer information about the objects and handles occlusions better.

The embedding step of the semantic labeling branch, as shown in Fig.~\ref{fig:network}, takes two feature maps with channel dimension 512 generated by the feature extraction stage as inputs. The resolutions of the two feature maps are $1/8$ and $1/16$ of the original image size, respectively. The network first reduces the channel dimension of the two feature maps to 64 using two convolutional layers. Then it doubles the resolution of the $1/16$ feature map with a deconvolutional layer. After that, the two feature maps are summed and another deconvolutional layer is used to increase the resolution by 8 times in order to obtain a feature map with the original image size. Finally, a convolutional layer operates on the feature map and generates semantic labeling scores for pixels. The output of this layer has $n$ channels with $n$ the number of the semantic classes. In training, a softmax cross entropy loss is applied to train the semantic labeling branch. While in testing, a softmax function is used to compute the class probabilities of the pixels. The design of the semantic labeling branch is inspired by the fully convolutional network in \cite{long2015fully} for semantic labeling. It is also used in our previous work for scene labeling \cite{xiang2017darnn}.

\subsection{3D Translation Estimation}

\begin{figure}
	\centering
	\includegraphics[height = 0.5\linewidth, width = \linewidth]{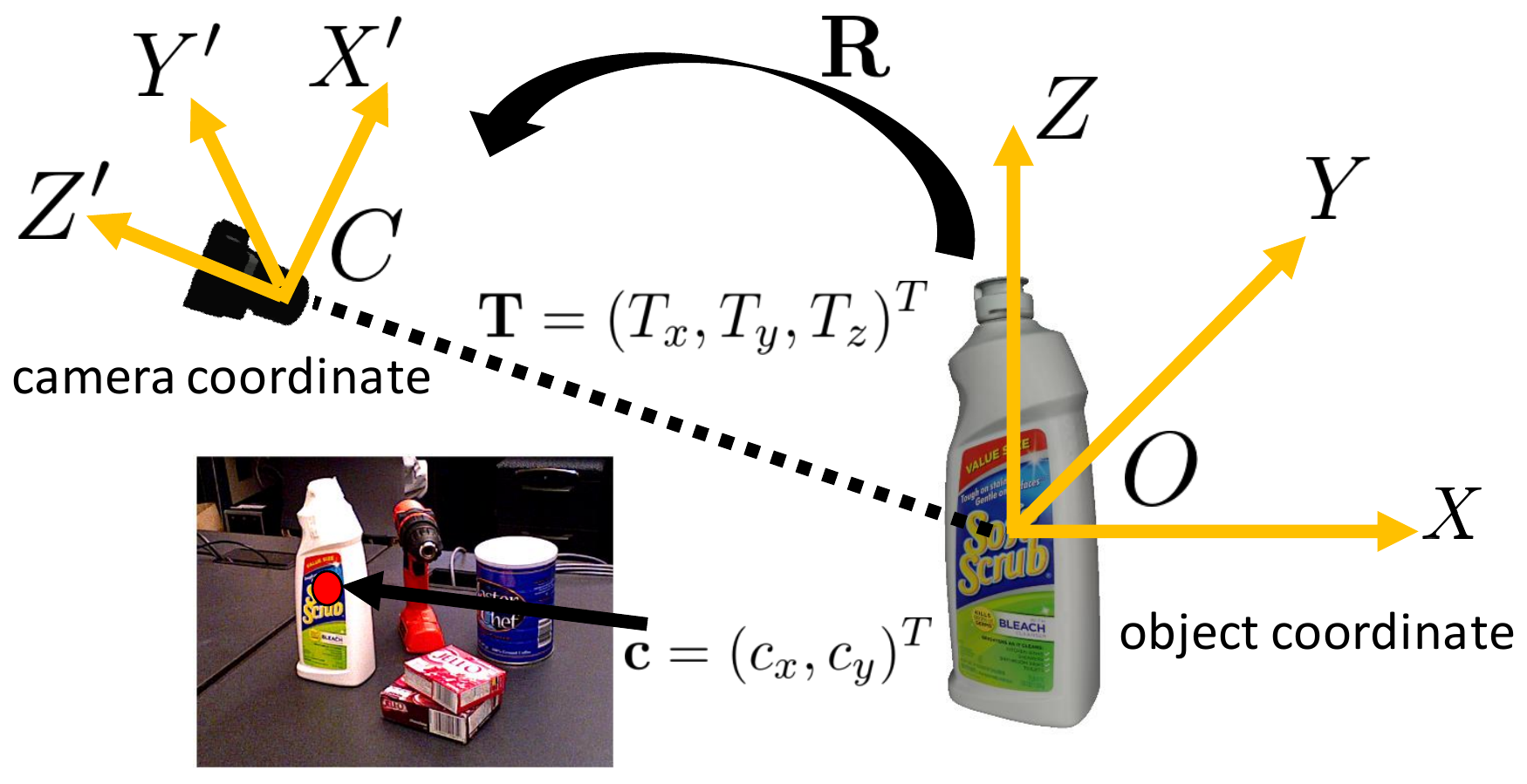}
	\caption{Illustration of the object coordinate system and the camera coordinate system. The 3D translation can be estimated by localizing the 2D center of the object and estimating the 3D center distance from the camera.}
	\label{fig:translation}
	\vspace{-6mm}
\end{figure}

% The next task is estimating the object center and its distance.

As illustrated in Fig.~\ref{fig:translation}, the 3D translation $\mathbf{T} = (T_x, T_y, T_z)^T$ is the coordinate of the object origin in the camera coordinate system. A naive way of estimating $\mathbf{T}$ is to directly regress the image features to $\mathbf{T}$. However, this approach is not generalizable since objects can appear in any location in the image. Also, it cannot handle multiple object instances in the same category. Therefore, we propose to estimate the 3D translation by localizing the 2D object center in the image and estimating object distance from the camera. To see, suppose the projection of $\mathbf{T}$ on the image is $\mathbf{c} = (c_x, c_y)^T$. If the network can localize $\mathbf{c}$ in the image and estimate the depth $T_z$, then we can recover $T_x$ and $T_y$ according to the following projection equation assuming a pinhole camera:
\begin{equation} \label{eq:projection}
\begin{bmatrix}
c_x \\[0.5em] c_y
\end{bmatrix} = \begin{bmatrix}
f_x \frac{T_x}{T_z} + p_x \\[0.5em]
f_y \frac{T_y}{T_z} + p_y
\end{bmatrix},
\end{equation}
where $f_x$ and $f_y$ denote the focal lengths of the camera, and $(p_x, p_y)^T$ is the principal point. If the object origin $O$ is the centroid of the object, we call $\mathbf{c}$ the 2D center of the object.

A straightforward way for localizing the 2D object center is to directly detect the center point as in existing key point detection methods \cite{pavlakos2017,cao2017realtime}. However, these methods would not work if the object center is occluded.  Inspired by the traditional Implicit Shape Model (ISM) in which image patches vote for the object center for detection \cite{leibe2004combined}, we design our network to regress to the center \emph{direction} for each pixel in the image. Specifically, for a pixel $\mathbf{p} = (x, y)^T$ on the image, it regresses to three variables:
\begin{equation} \label{eq::center}
(x, y) \rightarrow \Big(n_x = \frac{c_x - x}{\| \mathbf{c} - \mathbf{p} \|}, n_y = \frac{c_y - y}{\| \mathbf{c} - \mathbf{p} \|}, T_z\Big).
\end{equation}
Note that instead of directly regressing to the displacement vector $\mathbf{c} - \mathbf{p}$, we design the network to regress to the unit length vector $\mathbf{n} = (n_x, n_y)^T = \frac{\mathbf{c} - \mathbf{p}}{\| \mathbf{c} - \mathbf{p} \|}$, \emph{i.e}., 2D center direction, which is scale-invariant and therefore easier to be trained (as we verified experimentally).

The center regression branch of our network (Fig.~\ref{fig:network}) uses the same architecture as the semantic labeling branch, except that the channel dimensions of the convolutional layers and the deconvolutional layers are different. We embed the high-dimensional features into a 128-dimensional space instead of 64-dimensional since this branch needs to regress to three variables for each object class. The last convolutional layer in this branch has channel dimension $3 \times n$ with $n$ the number of object classes. In training, a smoothed L1 loss function is applied for regression as in \cite{girshick2015fast}.

\begin{figure}
	\centering
	\includegraphics[width = 0.6\linewidth]{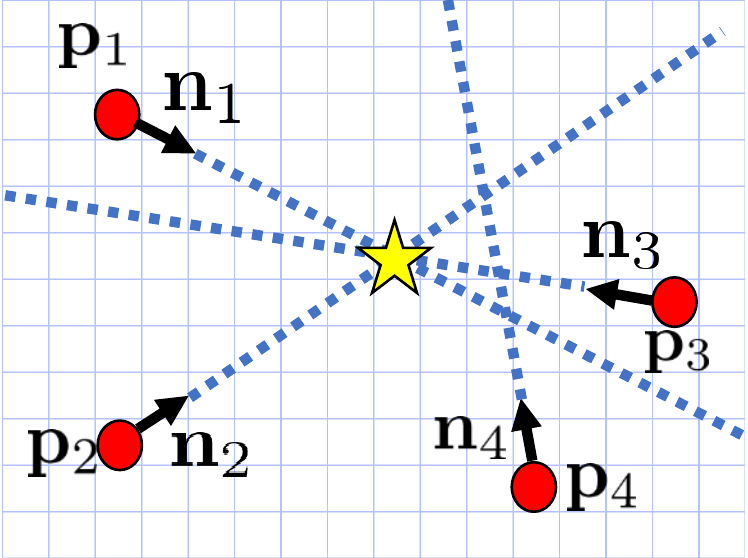}
	\caption{Illustration of Hough voting for object center localization: Each pixel casts votes for image locations along the ray predicted from the network.}
	\vspace{-4mm}
	\label{fig:hough}
\end{figure}

In order to find the 2D object center $\mathbf{c}$ of an object, a Hough voting layer is designed and integrated into the network. The Hough voting layer takes the pixel-wise semantic labeling results and the center regression results as inputs. For each object class, it first computes the voting score for every location in the image. The voting score indicates how likely the corresponding image location is the center of an object in the class. Specifically, each pixel in the object class adds votes for image locations along the ray predicted from the network (see Fig.~\ref{fig:hough}). After processing all the pixels in the object class, we obtain the voting scores for all the image locations. Then the object center is selected as the location with the maximum score. For cases where multiple instances of the same object class may appear in the image, we apply non-maximum suppression to the voting scores, and then select locations with scores larger than a certain threshold.

After generating a set of object centers, we consider the pixels that vote for an object center to be the inliers of the center. Then the depth prediction of the center, $T_z$, is simply computed as the mean of the depths predicted by the inliers. Finally, using Eq.~\ref{eq:projection}, we can estimate the 3D translation $\mathbf{T}$. In addition, the network generates the bounding box of the object as the 2D rectangle that bounds all the inliers, and the bounding box is used for 3D rotation regression.

\subsection{3D Rotation Regression}

The lowest part of Fig.~\ref{fig:network} shows the 3D rotation regression branch. Using the object bounding boxes predicted from the Hough voting layer, we utilize two RoI pooling layers~\cite{girshick2015fast} to ``crop and pool" the visual features generated by the first stage of the network for the 3D rotation regression. The pooled feature maps are added together and fed into three Fully-Connected (FC) layers. The first two FC layers have dimension 4096, and the last FC layer has dimension $4 \times n$ with $n$ the number of object classes. For each class, the last FC layer outputs a 3D rotation represented by a quaternion.

To train the quaternion regression, we propose two loss functions, one of which is specifically designed to handle symmetric objects.  The first loss, called PoseLoss (\pl), operates in the 3D model space and measures the average squared distance between points on the correct model pose and their corresponding points on the model using the estimated orientation. \pl\ is defined as
\begin{equation}
\text{\pl}(\mathbf{\tilde{q}}, \mathbf{q}) = \frac{1}{2m}\sum_{\mathbf{x} \in \mathcal{M}}\| R(\mathbf{\tilde{q}}) \mathbf{x} -  R(\mathbf{q})\mathbf{x}  \|^2,
\end{equation}
where $\mathcal{M}$ denotes the set of 3D model points and $m$ is the number of points. $R(\mathbf{\tilde{q}})$ and $R(\mathbf{q})$ indicate the rotation matrices computed from the the estimated quaternion and the ground truth quaternion, respectively. This loss has its unique minimum when the estimated orientation is identical to the ground truth orientation~\footnote{It is very similar to a regression loss on the quaternions, as we have verified experimentally. We use this formulation for consistency with the other loss.}.  Unfortunately, \pl\ does not handle symmetric objects appropriately, since a symmetric object can have \emph{multiple} correct 3D rotations. Using such a loss function on symmetric objects unnecessarily penalizes the network for regressing to one of the alternative 3D rotations, thereby giving possibly inconsistent training signals.

While \pl\ could potentially be modified to handle symmetric objects by manually specifying object symmetries and then considering all correct orientations as ground truth options, we here introduce ShapeMatch-Loss (\sloss), a loss function that does not require the specification of symmetries. \sloss\ is defined as
\begin{equation}
\text{\sloss}(\mathbf{\tilde{q}}, \mathbf{q}) = \frac{1}{2m}\sum_{\mathbf{x}_1 \in \mathcal{M}} \min_{\mathbf{x}_2 \in \mathcal{M}} \| R(\mathbf{\tilde{q}}) \mathbf{x}_1 - R(\mathbf{q}) \mathbf{x}_2  \|^2.
\end{equation}
As we can see, just like ICP, this loss measures the offset between each point on the estimated model orientation and the \emph{closest} point on the ground truth model. \sloss\ is minimized when the two 3D models match each other. In this way, the \sloss\ will not penalize rotations that are equivalent with respect to the 3D shape symmetry of the object.

\begin{figure}
	\centering
	\includegraphics[height = 0.8\linewidth, width = \linewidth]{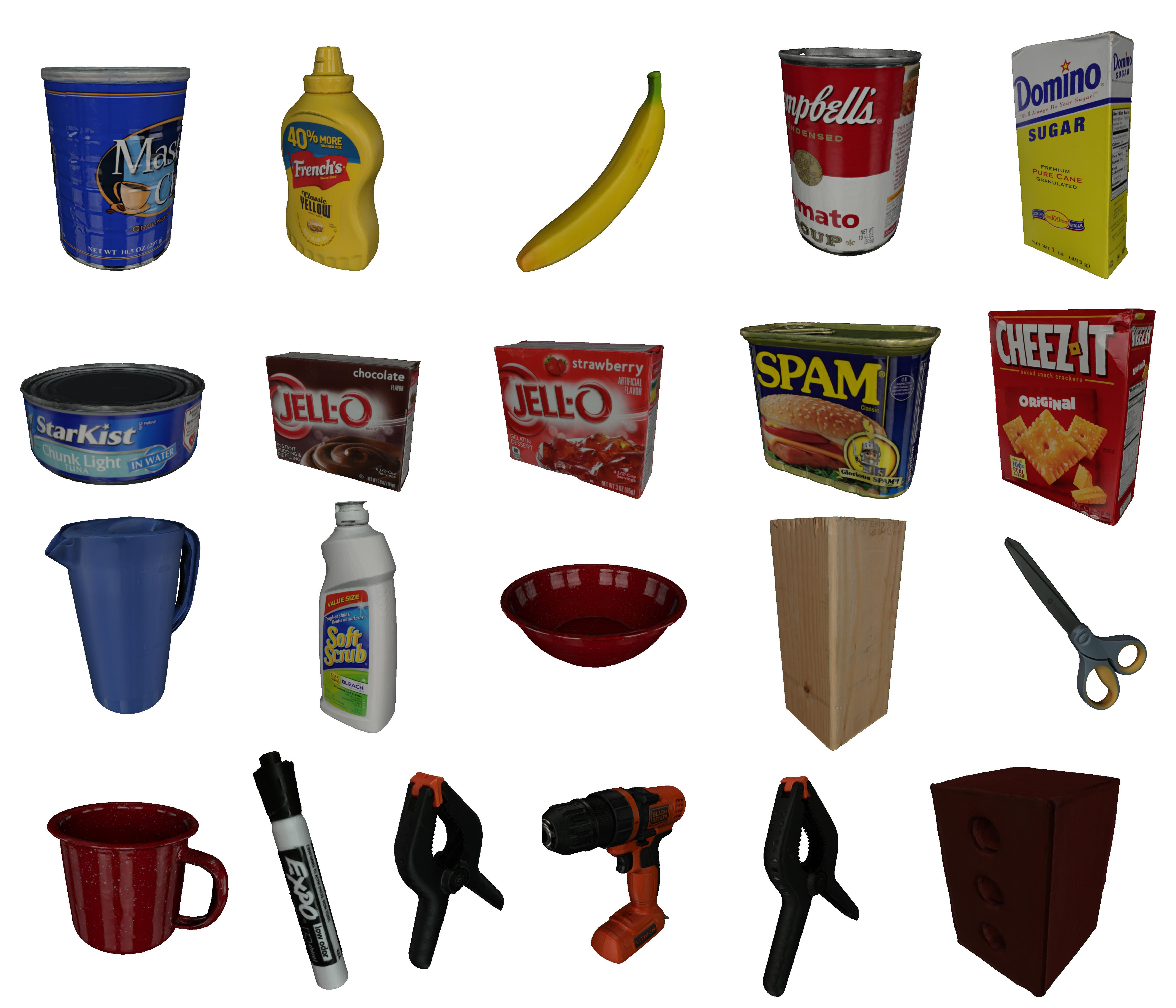}
	\caption{The subset of 21 YCB Objects selected to appear in our dataset.}
	\label{fig:ycb}
	\vspace{-2mm}
\end{figure}

\section{The YCB-Video Dataset}

Object-centric datasets providing ground-truth annotations for object poses and/or segmentations are limited in size by the fact that the annotations are typically provided manually. For example, the popular LINEMOD dataset \cite{hinterstoisser2012model} provides manual annotations for around 1,000 images for each of the 15 objects in the dataset. While such a dataset is useful for evaluation of model-based pose estimation techniques, it is orders of magnitude smaller than a typical dataset for training state-of-the-art deep neural networks. One solution to this problem is to augment the data with synthetic images. However, care must be taken to ensure that performance generalizes between real and rendered scenes.

\subsection{6D Pose Annotation}

To avoid annotating all the video frames manually, we manually specify the poses of the objects only in the first frame of each video.  Using Signed Distance Function (SDF) representations of each object, we refine the pose of each object in the first depth frame. Next, the camera trajectory is initialized by fixing the object poses relative to one another and tracking the object configuration through the depth video. Finally, the camera trajectory and relative object poses are refined in a global optimization step.

%Instead of annotating all the video frames manually, we first provide coarse manual annotations for the first frame of a collection of RGB-D videos, then use a model-based tracking system to estimate the camera trajectory and relative poses between objects. This allows us to compute the pose of each object in each individual frame by composing the object-to-world transform and the world-to-camera transform for the frame. A segmentation of each object is generated by rendering the 3D model according to its pose. In this way, each of a small number of manual annotations can be used by the tracker to generate thousands of labelled images. Furthermore, because the annotation for the first frame is not final but simply an initialization for the tracker, it does not need to be nearly as accurate and can therefore be generated much faster than a typical manual annotation.

\subsection{Dataset Characteristics}

\begin{table}
	\centering
	\caption{\small Statistics of our YCB-Video Dataset}
	\label{tab:DatasetStatistics}
	\begin{tabular}{ | c | c | }
		\hline			
		Number of Objects & 21 \\
		Total Number of Videos & 92 \\
		Held-out Videos & 12 \\
		Min Object Count & 3 \\
		Max Object Count & 9 \\
		Mean Object Count & 4.47 \\
		Number of Frames & 133,827 \\
		Resolution & 640 x 480 \\
		\hline
	\end{tabular}
	\vspace{-2mm}
\end{table}

\begin{figure}
	\centering
	\includegraphics[width=0.45\linewidth]{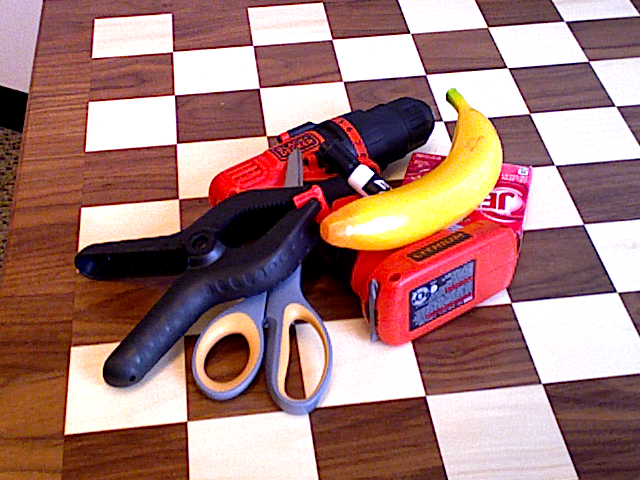}
	\includegraphics[width=0.45\linewidth]{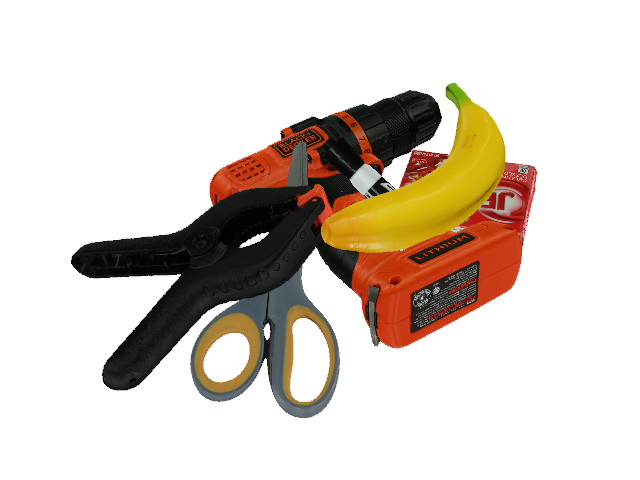}
	\caption{\small \textbf{Left:} an example image from the dataset. \textbf{Right:} Textured 3D object models (provided with the YCB dataset) rendered according to the pose annotations for this frame. }
	\label{fig:DatasetExample}
	\vspace{-4mm}
\end{figure}

The objects we used are a subset of 21 of the YCB objects \cite{calli2015ycb} as shown in Fig.~\ref{fig:ycb}, selected due to high-quality 3D models and good visibility in depth. The videos are collected using an Asus Xtion Pro Live RGB-D camera in fast-cropping mode, which provides RGB images at a resolution of 640x480 at 30 FPS by capturing a 1280x960 image locally on the device and transmitting only the center region over USB. This results in higher effective resolution of RGB images at the cost of a lower FOV, but given the minimum range of the depth sensor this was an acceptable trade-off. The full dataset comprises 133,827 images, two full orders of magnitude larger than the LINEMOD dataset. For more statistics relating to the dataset, see Table \ref{tab:DatasetStatistics}. Fig.~\ref{fig:DatasetExample} shows one annotation example in our dataset where we render the 3D models according to the annotated ground truth pose. Note that our annotation accuracy suffers from several sources of error, including the rolling shutter of the RGB sensor, inaccuracies in the object models, slight asynchrony between RGB and depth sensors, and uncertainties in the intrinsic and extrinsic parameters of the cameras.

%Fig.~\ref{fig:StereoExample} shows an example of a stereo pair generated from our video dataset.

%One drawback of our approach to dataset collection is that the accuracy of each annotation suffers from several sources of error, including the rolling shutter of the RGB sensor, inaccuracies in the object models, slight asynchrony between RGB and depth sensors, and uncertainties in the intrinsic and extrinsic parameters of the cameras. However, we hope that our dataset will still prove useful by making up for in scale where it lacks in accuracy. Another drawback is that the frames within one video are highly correlated, as they depict the same objects in a static configuration in a single scene.
%However, the motion of the camera through the video induces large appearance changes and leads to the observation of a variety of occlusions.
%As has been shown by the use of data augmentation by random cropping, images that appear nearly identical to a human can be independently valuable in training a deep neural network, and we believe the same effect is in play here.

%\begin{figure}
%	\centering
%	\includegraphics[width=0.45\linewidth]{left.png}
%	\includegraphics[width=0.45\linewidth]{right.png}
%	\caption{\small An example ``stereo'' pair, generated by pairing regularly spaced ``left'' frames with the first succeeding frame for which the tracked camera pose is at least \BaselineInCentimeters~cm distant.}
%	\label{fig:StereoExample}\vspace*{-3ex}
%\end{figure}

\section{EXPERIMENTS}

%In this section, we conduct experiments to evaluate our proposed method for 6D object pose estimation.

\subsection{Datasets}

In our YCB-Video dataset, we use 80 videos for training, and test on 2,949 key frames extracted from the rest 12 test videos. We also evaluate our method on the OccludedLINEMOD dataset \cite{krull2015learning}. The authors of \cite{krull2015learning} selected one video with 1,214 frames from the original LINEMOD dataset \cite{hinterstoisser2012model}, and annotated ground truth poses for eight objects in that video: Ape, Can, Cat, Driller, Duck, Eggbox, Glue and Holepuncher. There are significant occlusions between objects in this video sequence, which makes this dataset challenging. For training, we use the eight sequences from the original LINEMOD dataset corresponding to these eight objects. In addition, we generate 80,000 synthetic images for training on both datasets by randomly placing objects in a scene.

\subsection{Evaluation Metrics}

We adopt the average distance (ADD) metric as proposed in \cite{hinterstoisser2012model} for evaluation. Given the ground truth rotation $\mathbf{R}$ and translation $\mathbf{T}$ and the estimated rotation $\mathbf{\tilde{R}}$ and translation $\mathbf{\tilde{T}}$, the average distance computes the mean of the pairwise distances between the 3D model points transformed according to the ground truth pose and the estimated pose:
\begin{equation}
\textsc{ADD} = \frac{1}{m}\sum_{\mathbf{x} \in \mathcal{M}}\| (\mathbf{R} \mathbf{x} + \mathbf{T}) - (\mathbf{\tilde{R}} \mathbf{x} + \mathbf{\tilde{T}})  \|,
\end{equation}
where $\mathcal{M}$ denotes the set of 3D model points and $m$ is the number of points. The 6D pose is considered to be correct if the average distance is smaller than a predefined threshold. In the OccludedLINEMOD dataset, the threshold is set to 10\% of the 3D model diameter. For symmetric objects such as the Eggbox and Glue, the matching between points is ambiguous for some views. Therefore, the average distance is computed using the closest point distance:
\begin{equation} \label{eq::dsym}
\textsc{ADD-S} = \frac{1}{m}\sum_{\mathbf{x}_1 \in \mathcal{M}} \min_{\mathbf{x}_2 \in \mathcal{M}} \| (\mathbf{R} \mathbf{x}_1 + \mathbf{T}) - (\mathbf{\tilde{R}} \mathbf{x}_2 + \mathbf{\tilde{T}})  \|.
\end{equation}
Our design of the loss function for rotation regression is motivated by these two evaluation metrics. Using a fixed threshold in computing pose accuracy cannot reveal how a method performs on these incorrect poses with respect to that threshold. Therefore, we vary the distance threshold in evaluation. In this case, we can plot an accuracy-threshold curve, and compute the area under the curve for pose evaluation.

Instead of computing distances in the 3D space, we can project the transformed points onto the image, and then compute the pairwise distances in the image space. This metric is called the reprojection error that is widely used for 6D pose estimation when only color images are used.

\subsection{Implementation Details}

PoseCNN is implemented using the TensorFlow library \cite{abadi2016tensorflow}. The Hough voting layer is implemented on GPU as in \cite{van2011fast}. In training, the parameters of the first 13 convolutional layers in the feature extraction stage and the first two FC layers in the 3D rotation regression branch are initialized with the VGG16 network \cite{simonyan2014very} trained on ImageNet \cite{deng2009imagenet}. No gradient is back-propagated via the Hough voting layer. Stochastic Gradient Descent (SGD) with momentum is used for training.

\subsection{Baselines}

\textbf{3D object coordinate regression network.} Since the state-of-the-art 6D pose estimation methods mostly rely on regressing image pixels to 3D object coordinates \cite{brachmann2014learning,brachmann2016uncertainty,michel2016global}, we implement a variation of our network for 3D object coordinate regression for comparison. In this network, instead of regressing to center direction and depth as in Fig.~\ref{eq::center}, we regress each pixel to its 3D coordinate in the object coordinate system. We can use the same architecture since each pixel still regresses to three variables for each class. Then we remove the 3D rotation regression branch. Using the semantic labeling results and 3D object coordinate regression results, the 6D pose is recovered using the pre-emptive RANSAC as in \cite{brachmann2016uncertainty}.

\textbf{Pose refinement.} The 6D pose estimated from our network can be refined when depth is available. We use the Iterative Closest Point (ICP) algorithm to refine the 6D pose. Specifically, we employ ICP with projective data association and a point-plane residual term. We render a predicted point cloud given the 3D model and an estimated pose, and assume that each observed depth value is associated with the predicted depth value at the same pixel location. The residual for each pixel is then the smallest distance from the observed point in 3D to the plane defined by the rendered point in 3D and its normal. Points with residuals above a specified threshold are rejected and the remaining residuals are minimized using gradient descent. Semantic labels from the network are used to crop the observed points from the depth image. Since ICP is not robust to local minimums, we refinement multiple poses by perturbing the estimated pose from the network, and then select the best refined pose using the alignment metric proposed in \cite{wong2017segicp}.

% We experiment and evaluate different ways in refining the initial pose predicted from the network. When depth images are available, we refine the pose using the Iterative Closest Point (ICP) algorithm. When multi-view images are available, we triangulate the 2D object centers predicted from the network to refine the 3D translations, and also select the best 6D pose in a common 3D coordinate system for all the images as described in Sec.~\ref{sec:refinement}.

\subsection{Analysis on the Rotation Regress Losses}

\begin{figure}
	\centering
	\includegraphics[height = 0.85\linewidth, width = \linewidth]{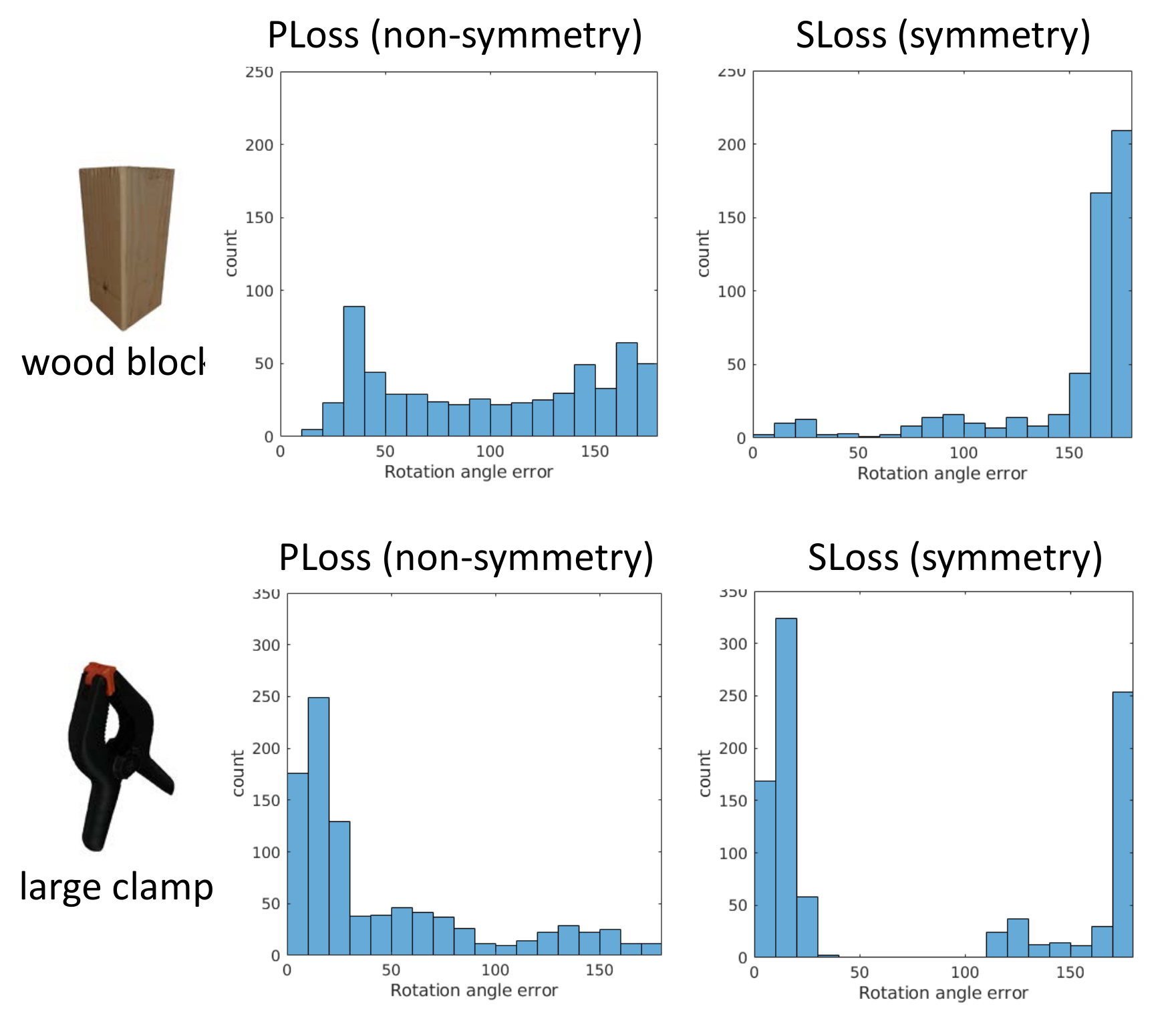}
	\caption{Comparison between the \pl\ and the \sloss\ for 6D pose estimation on three symmetric objects in the YCB-Video dataset.}
	\label{fig:loss}
	\vspace{-6mm}
\end{figure}

We first conduct experiments to analyze the effect of the two loss functions for rotation regression on symmetric objects. Fig.~\ref{fig:loss} shows the rotation error histograms for two symmetric objects in the YCB-Video dataset (wood block and large clamp) using the two loss functions in training. The rotation errors of the \pl\ for the wood block and the large clamp span from 0 degree to 180 degree. The two histograms indicate that the network is confused by the symmetric objects. While the histograms of the \sloss\ concentrate on the 180 degree error for the wood block and 0 degree and 180 degree for the large clamp, since they are symmetric with respect to 180 degree rotation around their coordinate axes.

\subsection{Results on the YCB-Video Dataset}

\begin{table*}
	\centering
	\caption{Area under the accuracy-threshold curve for 6D pose evaluation on the YCB-Video dataset. Red colored objects are symmetric.}
	\label{table:ycb}
	\begin{tabular}{|l|c|c|c|c||c|c|c|c|c|c|}
		\hline  & \multicolumn{4}{c||}{RGB} & \multicolumn{6}{c|}{RGB-D} \\
		\hline  & \multicolumn{2}{c|}{3D Coordinate}  & \multicolumn{2}{c||}{PoseCNN} & \multicolumn{2}{c|}{3D Coordinate} & \multicolumn{2}{c|}{3D Coordinate+ICP} & \multicolumn{2}{c|}{PoseCNN+ICP} \\
		\hline  Object & ADD & ADD-S & ADD & ADD-S & ADD & ADD-S & ADD  & ADD-S & ADD & ADD-S \\
		\hline 002\_master\_chef\_can & 12.3 & 34.4 &	\textbf{50.9} & \textbf{84.0} & 61.4 & 90.1  & \textbf{72.7} & 95.7 & 69.0 & \textbf{95.8} \\
		\hline 003\_cracker\_box & 16.8 & 40.0 & \textbf{51.7} & \textbf{76.9} & 57.4 & 77.4 & \textbf{82.7} & 91.0  & 80.7 & \textbf{91.8} \\
		\hline 004\_sugar\_box & 28.7 & 48.9 & \textbf{68.6} & \textbf{84.3} & 85.5 & 93.3 & 94.6 & 97.5 & \textbf{97.2} & \textbf{98.2} \\
		\hline 005\_tomato\_soup\_can & 27.3 & 42.2 &	\textbf{66.0} & \textbf{80.9} & 84.5 & 92.1 &	\textbf{86.1} & \textbf{94.5} & 81.6 & \textbf{94.5} \\
		\hline 006\_mustard\_bottle & 25.9 & 44.8 & \textbf{79.9} & \textbf{90.2} & 82.8 & 91.1 & \textbf{97.6} & 98.3 & 97.0 & \textbf{98.4} \\
		\hline 007\_tuna\_fish\_can & 5.4 & 10.4 & \textbf{70.4} & \textbf{87.9} & 68.8 & 86.9 & 76.7 & 91.4 & \textbf{83.1} & \textbf{97.1} \\
		\hline 008\_pudding\_box & 14.9 & 26.3 & \textbf{62.9} & \textbf{79.0} & 74.8 & 89.3 & 86.0 & 94.9 & \textbf{96.6} & \textbf{97.9} \\
		\hline 009\_gelatin\_box & 25.4 & 36.7 & \textbf{75.2} & \textbf{87.1} & 93.9 & 97.2 & \textbf{98.2} & \textbf{98.8} & \textbf{98.2} & \textbf{98.8} \\
		\hline 010\_potted\_meat\_can & 18.7 & 32.3 & \textbf{59.6} & \textbf{78.5} & 70.9 & 84.0 & 78.9 & 87.8 & \textbf{83.8} & \textbf{92.8} \\
		\hline 011\_banana & 3.2 & 8.8 & \textbf{72.3} & \textbf{85.9} & 50.7 & 77.3 & 73.5 & 94.3 & \textbf{91.6} & \textbf{96.9} \\
		\hline 019\_pitcher\_base & 27.3 & 54.3 & \textbf{52.5} & \textbf{76.8} & 58.2 & 83.8 & 81.1 & 95.6 & \textbf{96.7} & \textbf{97.8} \\
		\hline 021\_bleach\_cleanser & 25.2 & 44.3 & \textbf{50.5} & \textbf{71.9} & 74.1 & 89.2 & 87.2 & 95.7 & \textbf{92.3} & \textbf{96.8} \\
		\hline \textcolor{red}{024\_bowl} & 2.7 & 25.4 & \textbf{6.5} & \textbf{69.7} & 8.7 & 67.4 & 8.3 & 77.9 & \textbf{17.5} & \textbf{78.3} \\
		\hline 025\_mug & 9.0 & 20.0 & \textbf{57.7} & \textbf{78.0} & 57.1 & 85.3 & 67.0 & 91.1 & \textbf{81.4} & \textbf{95.1} \\
		\hline 035\_power\_drill & 18.0 & 36.1 & \textbf{55.1} & \textbf{72.8} & 79.4 & 89.4 & 93.2 & 96.2 & \textbf{96.9} & \textbf{98.0} \\
		\hline \textcolor{red}{036\_wood\_block} & 1.2 & 19.6 & \textbf{31.8} & \textbf{65.8} & 14.6 & 76.7 & 21.7 & 85.2 & \textbf{79.2} & \textbf{90.5} \\
		\hline 037\_scissors & 1.0 & 2.9 & \textbf{35.8} & \textbf{56.2} & 61.0 & 82.8 & 66.0 & 88.3 & \textbf{78.4} & \textbf{92.2} \\
		\hline 040\_large\_marker & 0.2 & 0.3 & \textbf{58.0} & \textbf{71.4} & 72.4 & 82.8 & 74.1 & 85.5 & \textbf{85.4} & \textbf{97.2} \\
		\hline \textcolor{red}{051\_large\_clamp} & 6.9 & 14.6 & \textbf{25.0} & \textbf{49.9} & 48.0 & 67.6 & \textbf{54.6} & 74.9 & 52.6 & \textbf{75.4} \\
		\hline \textcolor{red}{052\_extra\_large\_clamp} & 2.7 & 14.0 & \textbf{15.8} & \textbf{47.0} & 22.1 & 49.0 & 25.2 & 56.4 & \textbf{28.7} & \textbf{65.3} \\
		\hline \textcolor{red}{061\_foam\_brick} & 0.6 & 1.2 & \textbf{40.4} & \textbf{87.8} & 40.0 & 82.4 & 46.5 & 89.9 & \textbf{48.3} & \textbf{97.1} \\
		\hline
		\hline ALL & 15.1 & 29.8  & \textbf{53.7} & \textbf{75.9} & 64.6 & 83.7 & 74.5 & 90.1  & \textbf{79.3} & \textbf{93.0}  \\
		\hline
	\end{tabular}
	\vspace{-2mm}
\end{table*}

\begin{figure*}
	\centering
	\includegraphics[height = 0.28\linewidth, width = \linewidth]{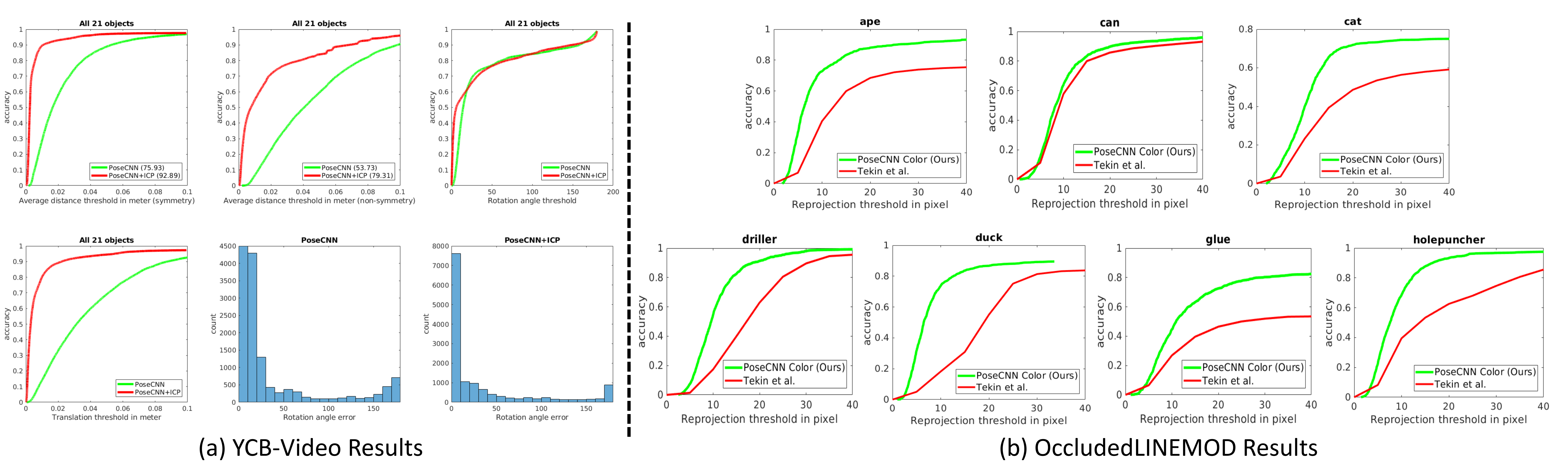}
	\caption{(a) Detailed results on the YCB-Video dataset. (b) Accuracy-threshold curves with reprojectin error on the OccludedLINEMOD dataset.}
	\label{fig::curves}
	\vspace{-6mm}
\end{figure*}

Table~\ref{table:ycb} and Fig. \ref{fig::curves}(a) presents detailed evaluation for all the 21 objects in the YCB-Video dataset. We show the area under the accuracy-threshold curve using both the ADD metric and the ADD-S metric, where we vary the threshold for the average distance and then compute the pose accuracy. The maximum threshold is set to 10cm.

We can see that i) By only using color images, our network significantly outperforms the 3D coordinate regression network combined with the pre-emptive RANSAC algorithm for 6D pose estimation. When there are errors in the 3D coordinate regression results, the estimated 6D pose can drift far away from the ground truth pose. While in our network, the center localization helps to constrain the 3D translation estimation even if the object is occluded. ii) Refining the poses with ICP significantly improves the performance. PoseCNN with ICP achieves superior performance compared to the 3D coordinate regression network when using depth images. The initial pose in ICP is critical for convergence. PoseCNN provides better initial 6D poses for ICP refinement. iii) We can see that some objects are more difficult to handle such as the tuna fish can that is small and with less texture. The network is also confused by the large clamp and the extra large clamp since they have the same appearance. The 3D coordinate regression network cannot handle symmetric objects very well such as the banana and the bowl.

\begin{figure*}
	\centering
	\includegraphics[height = 0.5\linewidth, width = 0.98\linewidth]{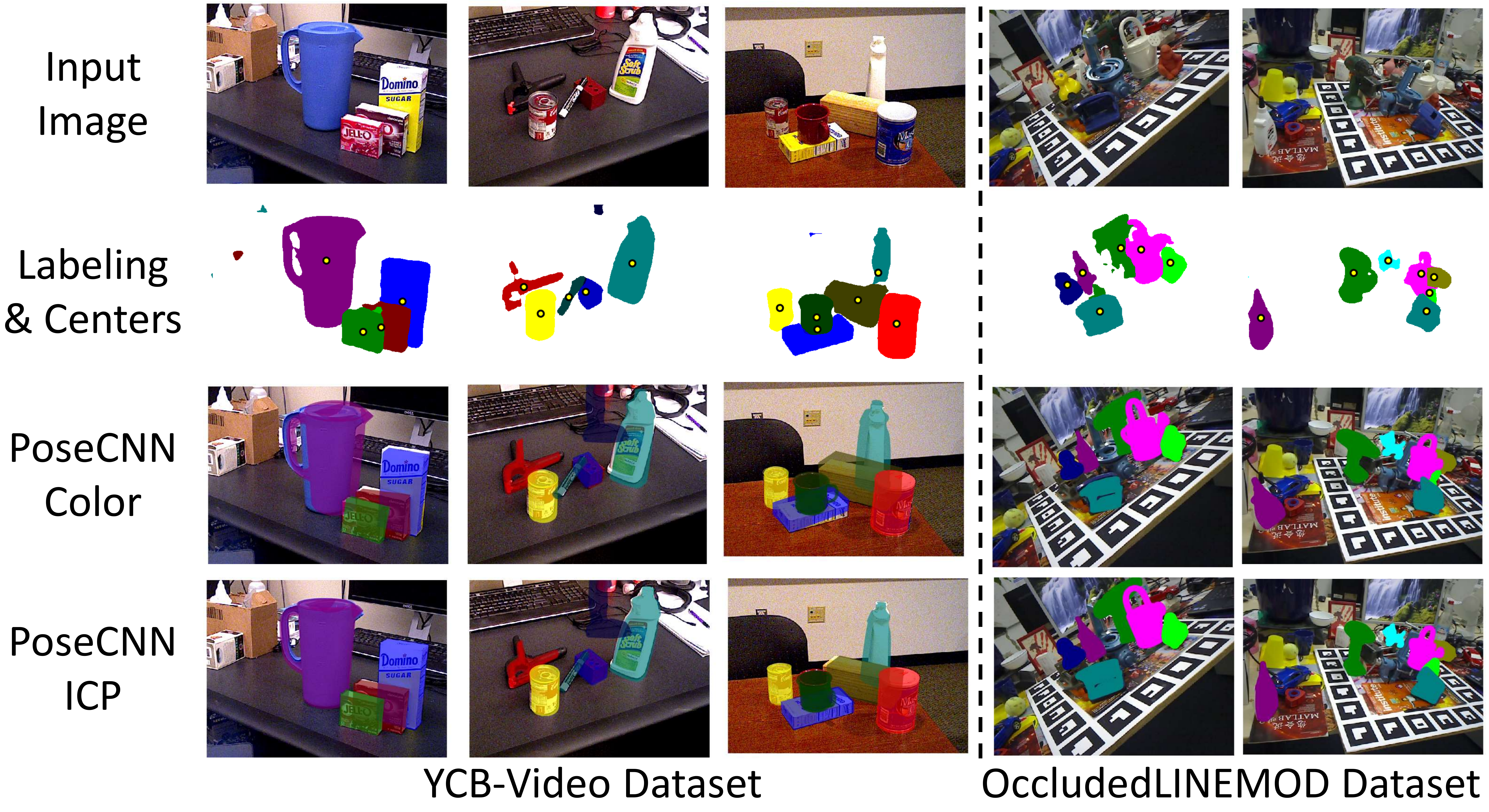}
	\caption{Examples of 6D object pose estimation results on the YCB-Video dataset from PoseCNN.}
	\label{fig::results}
	\vspace{-6mm}
\end{figure*}

Fig. \ref{fig::results} displays some 6D pose estimation results on the YCB-Video dataset. We can see that the center prediction is quite accurate even if the center is occluded by another object. Our network with color only is already able to provide good 6D pose estimation. With ICP refinement, the accuracy of the 6D pose is further improved.

\begin{table*} \setlength{\tabcolsep}{4pt}
	\centering
	\caption{6D pose estimation accuracy on the OccludedLINEMOD dataset. Red colored objects are symmetric. All methods use depth except for PoseCNN Color.}
	\label{table:linemod}
	\begin{tabular}{|l|c|c|c|c|c|c|c|}
		\hline Method & Michel et al. \cite{michel2016global}  & Hinterstoisser et al. \cite{hinterstoisser2016going} & Krull et al. \cite{krull2015learning} & Brachmann et al. \cite{brachmann2014learning} & Ours PoseCNN Color & Ours PoseCNN+ICP \\ \hline
		\hline Ape & 80.7 &	\textbf{81.4} & 68.0 & 53.1 & 9.6 & 76.2 \\
		\hline Can & 88.5 & \textbf{94.7} & 87.9 & 79.9 & 45.2 & 87.4 \\
		\hline Cat & \textbf{57.8} & 55.2 & 50.6 & 28.2 & 0.93 & 52.2 \\
		\hline Driller & \textbf{94.7} &	86.0 & 91.2 &	82.0 &  41.4 & 90.3 \\
		\hline Duck &	74.4 &	\textbf{79.7 }&	64.7 &	64.3 &	19.6 & 77.7 \\
		\hline \textcolor{red}{Eggbox} &	47.6 &	65.5 &	41.5 &	9.0 & 22.0 &	\textbf{72.2} \\
		\hline \textcolor{red}{Glue}	& 73.8 &	52.1 &	65.3 &	44.5 &	38.5 &	\textbf{76.7} \\
		\hline Holepuncher &	\textbf{96.3} &	95.5 &	92.9 &	91.6 &	22.1 &	91.4 \\
		\hline
		\hline MEAN &	76.7 &	76.3 &	70.3 &	56.6 &	24.9 & \textbf{78.0} \\
		\hline
	\end{tabular}
	\vspace{-2mm}
\end{table*}

\subsection{Results on the OccludedLINEMOD Dataset}

The OccludedLINEMOD dataset is challenging due to significant occlusions between objects. We first conduct experiments using color images only. Fig. \ref{fig::curves}(b) shows the accuracy-threshold curves with reprojection error for 7 objects in the dataset, where we compare PoseCNN with \cite{tekin2017real} that achieves the current state-of-the-art result on this dataset using color images as input. Our method significantly outperforms \cite{tekin2017real} by a large margin, especially when the reprojection error threshold is small. These results show that PoseCNN is able to correctly localize the target object even under severe occlusions.

By refining the poses using depth images in ICP, our method also outperforms the state-of-the-art methods using RGB-D data as input. Table~\ref{table:linemod} summarizes the pose estimation accuracy on the OccludedLINEMOD dataset. The most improvement comes from the two symmetric objects ``Eggbox'' and ``Glue''. By using our ShapeMatch-Loss for training, PoseCNN is able to correctly estimate the 6D pose of the two objects with respect to symmetry. We also present the result of PoseCNN using color only in Table~\ref{table:linemod}. These accuracies are much lower since the threshold here is usually smaller than 2cm. It is very challenging for color-based methods to obtain 6D poses within such small threshold when there are occlusions between objects. Fig. \ref{fig::results} shows two examples of the 6D pose estimation results on the OccludedLINEMOD dataset.

\section{CONCLUSIONS}

In this work, we introduce PoseCNN, a convolutional neural network for 6D object pose estimation. PoseCNN decouples the estimation of 3D rotation and 3D translation. It estimates the 3D translation by localizing the object center and predicting the center distance. By regressing each pixel to a unit vector towards the object center, the center can be estimated robustly independent of scale. More importantly, pixels vote the object center even if it is occluded by other objects. The 3D rotation is predicted by regressing to a quaternion representation. Two new loss functions are introduced for rotation estimation, with the ShapeMatch-Loss designed for symmetric objects. As a result, PoseCNN is able to handle occlusion and symmetric objects in cluttered scenes. We also introduce a large scale video dataset for 6D object pose estimation. Our results are extremely encouraging in that they indicate that it is feasible to accurately estimate the 6D pose of objects in cluttered scenes using vision data only. This opens the path to using cameras with resolution and field of view that goes far beyond currently used depth camera systems. We note that the \sloss\ sometimes results in local minimums in the pose space similar to ICP. It would be interesting to explore more efficient way in handle symmetric objects in 6D pose estimation in the future.

%% Use plainnat to work nicely with natbib.
\vspace{-2mm}
\section*{ACKNOWLEDGMENTS}

This work was funded in part by Siemens and by NSF STTR grant 63-5197 with Lula Robotics.

\bibliographystyle{plainnat}
\bibliography{references}

\end{document}